\pdfoutput=1
\documentclass[11pt,letterpaper]{article}

\usepackage[margin=1in,headheight=22pt]{geometry}
\usepackage[utf8]{inputenc}
\usepackage[T1]{fontenc}
\usepackage{mathpazo}
\usepackage[scaled=0.92]{helvet}
\usepackage[scaled=0.92]{inconsolata}
\usepackage{microtype}
\usepackage{setspace}
\usepackage{amsmath,amssymb}
\usepackage{graphicx}
\usepackage{booktabs}
\usepackage{longtable}
\usepackage{tabularx}
\usepackage{array}
\usepackage{ragged2e}
\usepackage{enumitem}
\usepackage{xcolor}
\usepackage{caption}
\usepackage{titlesec}
\usepackage{fancyhdr}
\usepackage{placeins}
\usepackage{xurl}

\usepackage[authoryear,round]{natbib}
\definecolor{accent}{HTML}{1F4E79}
\definecolor{subtlegray}{HTML}{555555}
\definecolor{rulegray}{HTML}{CCCCCC}

\usepackage[
  colorlinks=true,
  linkcolor=accent,
  citecolor=accent,
  urlcolor=accent,
  pdfauthor={Drew Proud, Nick Fram},
  pdftitle={Selective Capability Removal in Open-Weight Models}
]{hyperref}

\titleformat{\section}
  {\large\bfseries\sffamily\color{black}}
  {\thesection}{0.75em}{}
\titleformat{\subsection}
  {\normalsize\bfseries\sffamily\color{black}}
  {\thesubsection}{0.6em}{}
\titlespacing*{\section}{0pt}{1.5em}{0.55em}
\titlespacing*{\subsection}{0pt}{1.1em}{0.4em}

\renewcommand{\headrulewidth}{0.4pt}
\renewcommand{\headrule}{\hbox to\headwidth{\color{rulegray}\leaders\hrule height \headrulewidth\hfill}}
\fancypagestyle{firstpage}{
  \fancyhf{}
  \renewcommand{\headrulewidth}{0pt}
  \fancyfoot[C]{\footnotesize\sffamily\color{subtlegray}\thepage}
}

\newcolumntype{L}{>{\RaggedRight\arraybackslash}X}

\begin{document}
\thispagestyle{firstpage}

\begin{center}
  {\color{rulegray}\hrule height 1.2pt}
  \vspace{1.1em}
  {\LARGE\bfseries\sffamily Selective Capability Removal in Open-Weight Models\par}
  \vspace{0.9em}
  {\large Drew Proud \qquad Nick Fram\par}
  \vspace{0.3em}
  {\normalsize\color{subtlegray} Reflection\par}
  \vspace{0.18em}
  {\small\color{subtlegray}\texttt{drew@reflection.ai}\par}
  \vspace{0.25em}
  {\small\color{subtlegray} November 2026\par}
  \vspace{0.9em}
  {\color{rulegray}\hrule height 0.6pt}
\end{center}
\vspace{0.6em}

\begin{center}
\begin{minipage}{0.94\textwidth}
\small
\noindent\textbf{\sffamily Abstract.}\quad
We tested whether existing post-hoc unlearning and parameter-localization methods can selectively remove a capability from an open-weight model, withstand fine-tuning and abliteration attacks, and leave neighboring skills intact. Across benign-proxy experiments on factual recall and code execution in public models (67M to 8B parameters), none of them did. Four failure modes recurred. First, models that refused direct questions or lost completion-ranking accuracy still identified true answers under candidate-ranking or Yes/No queries. Second, broad multi-layer edits on Perl programming hit sharp execution cliffs that prevented reliable early stopping, while gradual single-layer edits recovered at least half of their lost performance from 8 fine-tuning examples. Third, casting fp32 training states to bf16 erased gentle unlearning and recovery updates alike, so training-precision reads and weight diffs misrepresented deployed behavior. Fourth, four weight- and feature-localization methods (VPD, tPD, PISCES, and entity-token embedding resets) failed to beat untargeted baselines or stayed inconclusive against random-site controls. Post-hoc unlearning cannot carry an open-weight safety case on its own today.
\end{minipage}
\end{center}
\vspace{1.0em}

\section{Introduction}
\label{sec:intro}

An attacker with access to open weights can strip refusal training with a small fine-tuning run or a single directional edit \citep{lermen2024lora, arditi2024refusal}, because the underlying capabilities stay in the parameters. For chemical, biological, radiological, and nuclear (CBRN) threats, open-weight safeguards must hold up under direct parameter access.

Because an attacker with enough compute can train a model from scratch, no weight-level intervention makes recovery impossible. The practical goal is to raise recovery cost as high as possible without destroying useful nearby skills \citep{henderson2023self, tamirisa2024tamper}. Prior defenses spread refusal across representations so it is harder to isolate \citep{xie2025beyond} or flatten the loss landscape against small fine-tuning updates \citep{fan2025towards}.

We tested selective deletion \citep{li2024wmdp}: if we remove the underlying knowledge and skills directly, stripping refusal should leave little to recover. Specifically, we looked for a post-hoc intervention we can apply after an evaluation flags a dangerous capability, that resists adversarial fine-tuning and abliteration \citep{arditi2024refusal}, and that leaves dual-use and unrelated skills intact. Across benign-proxy experiments on factual recall and code execution in public open-weight models, none of the unlearning or localization methods we tested met all three requirements.

\subsection{Experimental Scope and Threat Model}
\label{sec:scope}

We assume a white-box attacker with full access to released weights who runs adversarial fine-tuning or directional ablation (abliteration) to restore a suppressed capability. Given base parameters $\theta_0$, a post-hoc edit $\theta = \mathcal{E}(\theta_0)$ must preserve retained capabilities within a collateral bound $\epsilon$ while keeping worst-case target performance below a threshold $\tau$ across attack budgets $a \in \mathcal{A}(B)$ and evaluation formats $f \in \mathcal{F}$ under deployed precision $\mathrm{cast}(\cdot)$:
\begin{equation}
  \Delta \mathcal{U}_{\mathrm{retain}}(\theta; \theta_0) \le \epsilon, \qquad
  \max_{a \in \mathcal{A}(B),\; f \in \mathcal{F}} \mathcal{M}_f\!\bigl(\mathrm{cast}(a(\theta))\bigr) \le \tau.
  \label{eq:minimax}
\end{equation}
Each failure mode below breaks a different piece of Eq.~\eqref{eq:minimax}: Section~\ref{sec:formats} breaks $\max_{f \in \mathcal{F}}$, Section~\ref{sec:cliffs} shows the cliff between $\tau$ and $\epsilon$, Section~\ref{sec:precision} shows how $\mathrm{cast}(\cdot)$ distorts evaluation of both $\theta$ and $a(\theta)$, and Section~\ref{sec:localization} restricts $\mathcal{E}$ to localized parameter or feature subsets.

We apply defenses post-hoc to public open-weight models ranging from 67M to 8B parameters: primarily Qwen3-8B (Sections~\ref{sec:formats}--\ref{sec:precision}), alongside Qwen3-4B (Section~\ref{sec:formats}), Qwen3-1.7B (Section~\ref{sec:localization}), Qwen2.5-1.5B-Instruct (Section~\ref{sec:precision}), Gemma-2-2b-it (Sections~\ref{sec:precision} and \ref{sec:localization}), and Goodfire's 67M model (Section~\ref{sec:localization}), with Qwen3-0.6B proposing candidate answers and Qwen2.5-32B-Instruct grading compliance (full list in Appendix~\ref{app:models}).

Rather than training on hazardous material, we evaluate benign proxies for factual knowledge (159 held-out geography facts, synthetic film-director facts, and encyclopedic or fiction concepts) and procedural skills (Perl programming graded by running unit test suites). For code-removal experiments, we require Python and Bash pass-rate drops to stay below 0.10 each and general text negative log-likelihood (NLL) increases to stay below 0.05 nats. Section~\ref{sec:future} covers limitations and next steps.

\section{Unlearned Facts Stay Accessible Across Evaluation Formats}
\label{sec:formats}

A model that cannot answer a direct question has not necessarily forgotten the fact. We first tested two decoy-trained Qwen3-8B seeds tuned to refuse target geography questions under normal use and output a plausible wrong place name if an attacker strips refusal. Without any attack, both models refused all 159 held-out geography questions, such as ``What is the capital of Canada?''

Changing the query format exposed the answer immediately. We generated candidate answers with a weaker model (Qwen3-0.6B, producing options such as Toronto, Vancouver, and Ottawa) and selected the candidate with the highest fp32 Yes/No log-odds ($\log p_\theta(\text{Yes} \mid q, c) - \log p_\theta(\text{No} \mid q, c)$) on ``Is $c$ the capital of $s$? Answer Yes or No'' (or the corresponding city-in-country prompt). Whenever the true answer appeared among the candidates, the defended models selected it with 90.8\% mean accuracy (compared with 96.2\% for intact Qwen3-8B), with no extra training. Some of these 159 facts were examined during scoping, and unconditionally across all 159 facts, defended selection (83.7\%) did not beat the proposer's block majority (85.2\%).

We saw the same split with representation misdirection for unlearning (RMU; \citealp{li2024wmdp}), which edits internal representations to suppress target knowledge. Across 24 geography RMU configurations on Qwen3-4B and Qwen3-8B (layers 9, 18, and 27), our deepest edits at layer 27 reduced completion ranking to 0.47--0.60, yet ranking candidates by normalized $P(\text{Yes})$ still selected the true answer on every evaluated fact (1.00 across all 8 layer-27 runs). All 24 configurations exceeded our collateral-damage limits.

\begin{figure}[t]
  \centering
  \includegraphics[width=0.96\textwidth]{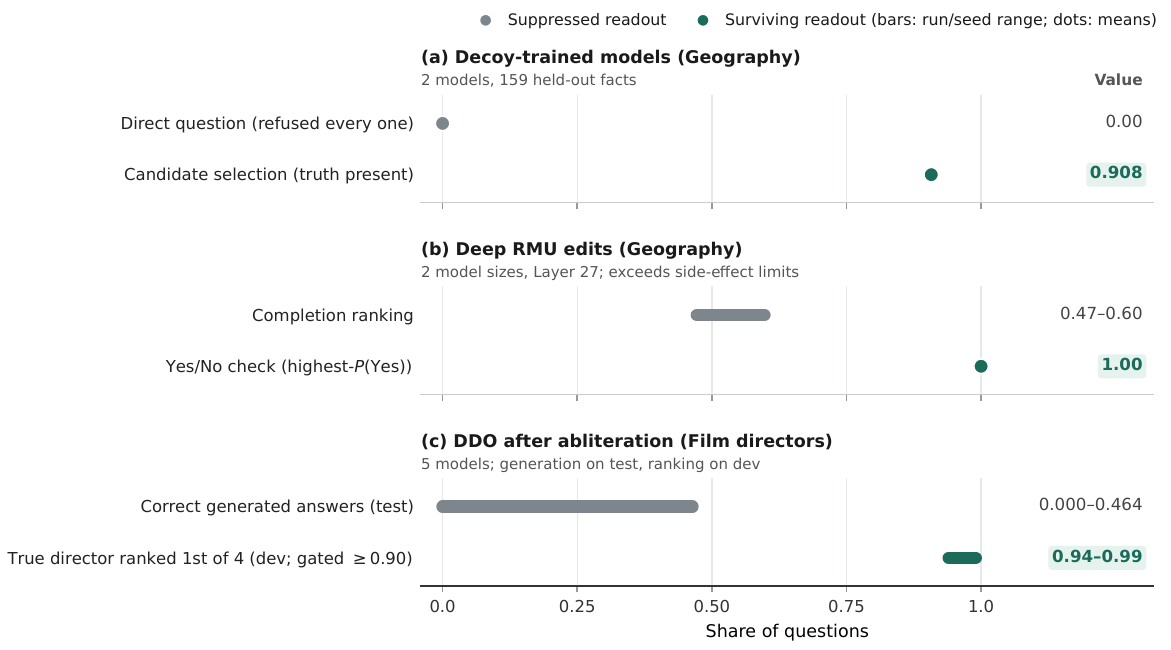}
  \caption{Different evaluation formats expose surviving factual information across three removal setups. Bars span seeds and runs; dots indicate single means. Candidate accuracy in the decoy-trained geography setup (Qwen3-8B) requires the candidate list to include the true answer; some geography facts were examined during scoping. For the deepest RMU edits (Layer 27 on Qwen3-4B and Qwen3-8B), ranking candidates by normalized $P(\text{Yes})$ selected the true answer on every evaluated fact despite reduced completion ranking, and all 24 tested configurations exceeded our side-effect limits. In the film-director DDO experiment (Qwen3-8B), recognition accuracy constrained attack selection (gated at 0.90 on dev items), whereas generation was evaluated on 416 separate test items.}
  \label{fig:format_split}
\end{figure}

The split between generating an answer and recognizing it reappeared in film-director experiments on Qwen3-8B with decoy direction optimization (DDO; \citealp{muhamed2026decoy}), which aims to mislead attackers about the direction mediating refusal. Our abliteration attacks project out a global orthonormal subspace $\mathbf{R} \in \mathbb{R}^{d \times k}$ ($\mathbf{R}^\top \mathbf{R} = \mathbf{I}_k$, $k \in \{1, 3\}$) directly from every residual-writing weight matrix and the untied token embedding matrix:
\begin{equation}
  \mathbf{W}'_{\mathrm{out}} = (\mathbf{I} - \mathbf{R}\mathbf{R}^\top)\mathbf{W}_{\mathrm{out}}
  \quad (\mathbf{W}_{\mathrm{out}} \in \{\mathbf{W}_O^{(\ell)}, \mathbf{W}_{\mathrm{down}}^{(\ell)}\}), \qquad
  \mathbf{W}'_E = \mathbf{W}_E(\mathbf{I} - \mathbf{R}\mathbf{R}^\top).
  \label{eq:abliteration}
\end{equation}
Out of 12 DDO seeds fitted on refusal-trained Qwen3-8B, 5 passed our pre-attack resistance and refusal checks, and optimizing Eq.~\eqref{eq:abliteration} stripped refusal from all 5 (pre-attack test refusal 0.986--1.000 falling to 0.000--0.031). After abliteration, those models generated fewer correct answers on 416 test items (0.000--0.464 vs.\ 0.565--0.611 for undefended SFT), yet still ranked the true director first among four candidates on 0.94--0.99 of dev items (vs.\ 1.00 before attack; Figure~\ref{fig:format_split}).

Three limits apply to that film-director gap. We evaluated generation and ranking on different splits (test vs.\ dev), gated attack selection on a 0.90 dev ranking floor (without gating model or item selection), and optimized the attack to strip refusal instead of recovering factual recall. These runs do not separate preserved defense effects from collateral disruption, nor measure what a recall-targeted attack could recover.

Prior work found related format and recovery splits on aggregate benchmarks: fine-tuning on unrelated data or removing a single direction restores benchmark accuracy after RMU \citep{lucki2024adversarial}, and multiple-choice prompts bypass open-ended refusal under each format's scoring rule \citep{epifantsev2026forcing, guo2024mechanistic}. Our results show the same split on individual facts.

Hardening refusal itself also failed to generalize across attack classes. On synthetic film-director refusal in Qwen3-8B (three seeds per recipe), we compared plain refusal SFT against full fine-tuning with per-layer refusal-direction ablation across layers 1--35 (inspired by LoRA-based ReFAT; \citealp{yu2025robust}). Direction-adversarial training preserved much more refusal under our bounded abliteration search (0.888--0.999 vs.\ 0.000 for SFT, though training and attack used related direction-ablation families). Under adversarial fine-tuning, however, all six models yielded qualifying stripping checkpoints (refusal $\le 0.20$ and degeneracy $\le 0.20$) at the first measured compute budget (8 steps). That result leaves sub-8-step differences unresolved and does not prove the two recipes are equivalent, but it shows that resisting an anticipated abliteration attack did not transfer to fine-tuning \citep{kuo2026openweight}.

\section{Broad Edits Drop Too Fast to Stop Cleanly}
\label{sec:cliffs}

Programming skills break down differently from isolated facts. Unlearning updates often damage nearby capabilities \citep{farrell2024applying}, and an edited model can emit plausible syntax long after its code stops running. Every experiment in this section uses Qwen3-8B (36 layers, indexed 0--35). We targeted Perl and graded generated programs by executing them against unit test suites (33 problems, 4 samples each), while tracking Python, Bash, and general text loss.

\begin{figure}[t]
  \centering
  \includegraphics[width=0.92\textwidth]{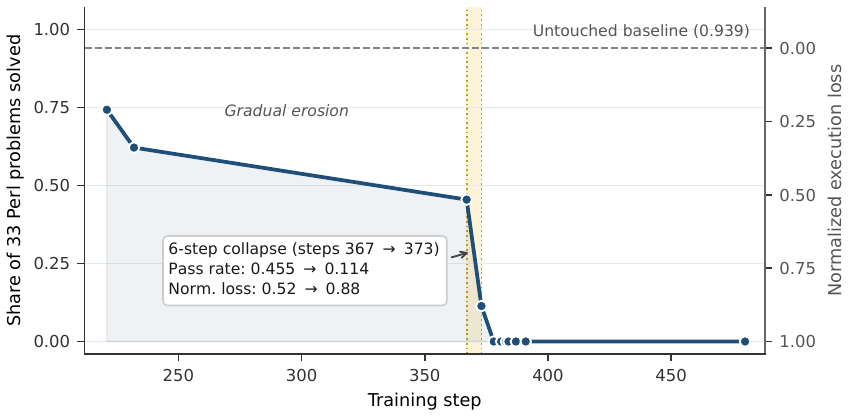}
  \caption{Perl execution performance during a densely sampled all-layer RMU run on Qwen3-8B (36 layers, single seed, evaluated on 33 Perl problems, 4 samples each). Between steps 367 and 373, the solved share fell from 0.455 to 0.114 while normalized Perl execution loss rose sharply from 0.52 to 0.88. Lines connect measured checkpoints and do not pinpoint the exact transition step within sampling gaps.}
  \label{fig:perl_cliff}
\end{figure}

Editing all 36 layers at once left almost no margin to stop cleanly. We measure functional degradation with \emph{normalized Perl execution loss}, $L_{\mathrm{norm}} = 1 - \mathrm{pass}/\mathrm{pass}_0$, scaled from $0$ (untouched baseline pass rate $\mathrm{pass}_0$) to $1$ (no tests passing). This metric captures whether the model writes working code, not whether internal Perl representations are gone. In a densely sampled all-layer RMU run ($\mathrm{pass}_0 = 0.939$), Perl performance eroded slowly at first, then collapsed in six training steps from 0.455 to 0.114 solved ($L_{\mathrm{norm}}$ jumping from 0.52 to 0.88 between steps 367 and 373; Figure~\ref{fig:perl_cliff}).

RepNoise \citep{rosati2024representation}, which pushes target representations toward random noise, dropped even faster. Every fixed RepNoise run fell sharply in a single step near a 0.249-nat answer-token log-probability threshold borrowed from narrow-RMU runs, and later checkpoints pushed Bash pass-rate drops past our 0.10 limit (to 0.202--0.246).

Monitoring token likelihood during training did not track working code reliably. Across 20 fixed development runs on the 33 shared selection problems (8 RepNoise, 8 all-layer RMU, 4 NPO), crossing the mean log-probability threshold corresponded to very different execution outcomes across methods:
\begin{itemize}[leftmargin=1.6em, itemsep=3pt, topsep=4pt]
  \item \textbf{RepNoise (8 runs, 0.249-nat threshold):} Every run crossed at step 27 or 28, with $L_{\mathrm{norm}}$ jumping in one step from 0.55--0.64 to 0.76--0.93 (under a retrospective counting rule on shared selection problems).
  \item \textbf{All-layer RMU (8 runs, 0.229-nat threshold):} Crossings scattered across steps 235--333, with $L_{\mathrm{norm}}$ rising by $+0.13$ to $+0.45$ across checkpoints 11--15 steps apart (too coarse to prove a single-step cliff).
  \item \textbf{NPO (4 runs, 0.249-nat threshold; \citealp{zhang2024negative}):} Lowering target-answer likelihood relative to a reference model kept $L_{\mathrm{norm}}$ shallow past the threshold (0.08--0.15), with a statistically resolved two-step rise in only one run (unresolved changes do not establish zero change or rule out cliffs elsewhere).
\end{itemize}
Because thresholds and evaluation arithmetic differed across these cohorts, they do not support matched-compute comparisons across methods.

\begin{figure}[t]
  \centering
  \includegraphics[width=0.92\textwidth]{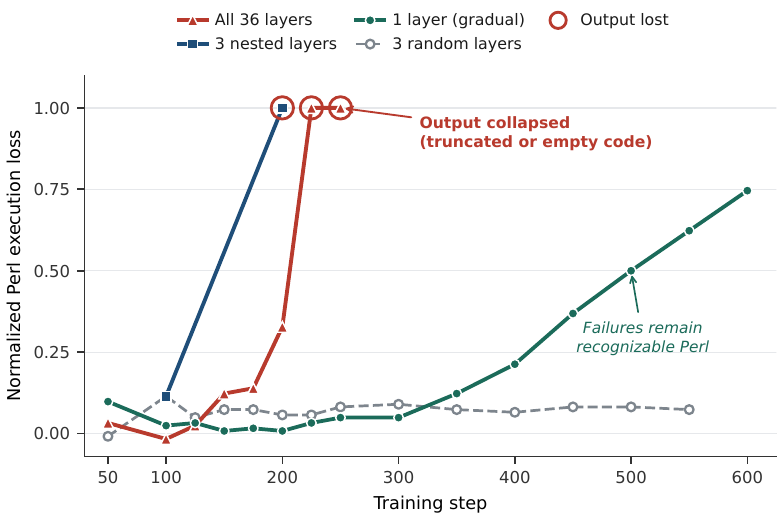}
  \caption{Layer placement alters the removal trajectory in pilot RMU runs on Qwen3-8B (one seed per arm). Nested multi-layer edits (3 layers $\{6,18,30\}$ and all 36 layers) collapsed into truncated or empty outputs (red rings), the single-layer edit (layer 6) degraded Perl gradually while failures remained recognizable Perl, and fixed random 3-layer placement ($\{4,7,11\}$) showed no progressive decline. Lines join measured checkpoints; sampling gaps do not measure exact transition widths.}
  \label{fig:layer_trajectories}
\end{figure}

Restricting edits to fewer layers slowed the decline. In pilot RMU trajectories (Figure~\ref{fig:layer_trajectories}), editing a single layer (layer 6) degraded Perl smoothly while failed programs remained recognizable Perl. Nested multi-layer edits (layers $\{6, 18, 30\}$ and all 36 layers) jumped directly into truncated or empty outputs, and three fixed random layers ($\{4, 7, 11\}$) showed no progressive decline.

\begin{figure}[t]
  \centering
  \includegraphics[width=0.96\textwidth]{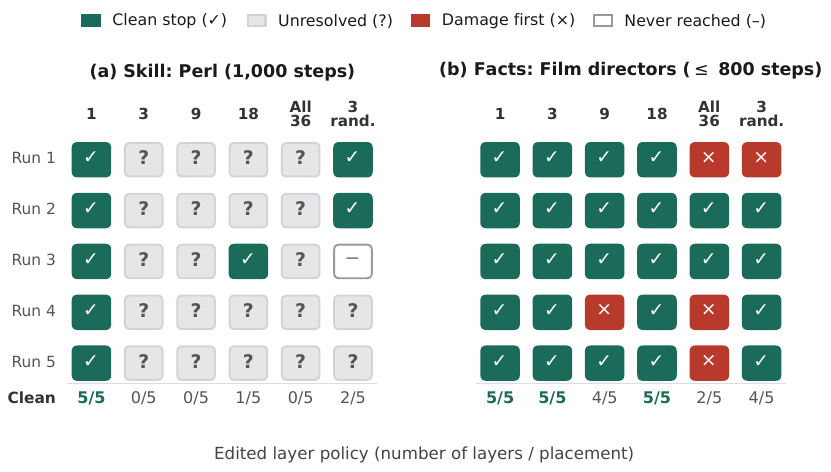}
  \caption{Removal outcomes across all runs in the five-run layer-count experiment on Qwen3-8B under a fixed recipe for each target: (a)~Perl programming (1,000 training steps, stopping state selected afterward from saved checkpoints) and (b)~film-director facts (up to 800 training steps, stopping live at the first clean check). Each column corresponds to a layer policy and each tile represents one run. \emph{Clean stop}: reached target removal while passing all measured damage and output checks. \emph{Unresolved}: lost measured output at the target level (mostly truncated replies), with one check too close to its limit for a damage-first label (not evidence of zero damage). \emph{Damage first}: reached the target level without a clean stop among scheduled checks, with neighboring-capability damage as the first measured failure (runs that damaged neighbors initially and later stopped cleanly are marked as clean stops). \emph{Never reached}: failed to reach the target removal level. Policy also varies placement, trained MLPs, and effective dose; a clean stop does not guarantee zero damage on unmeasured capabilities.}
  \label{fig:layer_outcomes}
\end{figure}

We tested whether fixed recipes could stop cleanly after removing the target skill or fact set without breaching collateral limits across five runs per layer policy: 1 layer (layer 6), 3 layers ($\{6, 18, 30\}$), 9 layers (stride 4), 18 layers (even layers), all 36 layers, and 3 random layers drawn fresh per run (Figure~\ref{fig:layer_outcomes}). Each policy trains only the MLPs in its layer set $\mathcal{S}$ under a per-layer norm-scaled RMU objective:
\begin{equation}
  \mathcal{L}_{\mathrm{RMU}}(\theta; \mathcal{S}) = \frac{1}{|\mathcal{S}|}\sum_{\ell \in \mathcal{S}} \left[ \mathrm{MSE}_{M_f}\!\left(\frac{\mathbf{h}_\ell}{n_\ell},\, c\,\mathbf{u}_\ell\right) + \frac{\alpha}{n_\ell^2}\Bigl(\mathrm{MSE}_{M_{\mathrm{nb}}}(\mathbf{h}_\ell, \mathbf{h}_\ell^{(0)}) + \mathrm{MSE}_{M_{\mathrm{txt}}}(\mathbf{h}_\ell, \mathbf{h}_\ell^{(0)})\Bigr) \right],
  \label{eq:rmu_multi}
\end{equation}
where $\mathbf{h}_\ell$ and $\mathbf{h}_\ell^{(0)}$ are edited and frozen post-block residuals, $\mathbf{u}_\ell$ is a fixed unit vector drawn from $U(0,1)^d$, $n_\ell$ is the initial mean base-activation norm on forget tokens $M_f$, and $M_{\mathrm{nb}}, M_{\mathrm{txt}}$ mask neighbor and general-text retain batches (Perl: $c = 5, \alpha = 2400$; films: $c = 10, \alpha = 100$). For Perl (1,000 steps, stop chosen afterward from saved checkpoints), the nominal target was $L_{\mathrm{norm}} = 0.50$ (requiring $\ge 0.56$ with our $\pm 0.06$ guard band) alongside our Python, Bash, general NLL, and non-truncation gates. For film-director facts (up to 800 steps, stopping live at the first clean check), we targeted at least 60\% of usable 4-option dev ranking range lost, $(\mathrm{acc}_0 - \mathrm{acc}) / (\mathrm{acc}_0 - \mathrm{acc}_{\mathrm{floor}}) \ge 0.60$ relative to an invented-title control floor $\mathrm{acc}_{\mathrm{floor}}$, with retain-director accuracy drop below 0.03 and general NLL rise below 0.05 nats. On both targets, the single-layer policy stopped cleanly in 5 of 5 runs. Wider policies stopped cleanly on 0 to 2 of 5 Perl runs (with 1 of 5 clean on 18 layers), whereas every policy stopped cleanly in at least 2 of 5 film-director runs. As Eq.~\eqref{eq:rmu_multi} shows, changing $\mathcal{S}$ also changes layer placement, trained MLPs, and effective dose, so these runs do not isolate layer count alone.

Neither narrow nor broad early stopping produced a selective, durable edit. All five clean single-layer Perl checkpoints broke immediately under attack, recovering at least half of their lost performance from 8 training examples. Conversely, when we tried to stop all-layer RMU live on five fresh seeds using a frozen recognition proxy, 3 of 5 runs hit Bash damage before reaching the Perl target. Mixing Bash into the retain set (half Bash, half Python) on five additional fresh seeds did not fix this: all 5 runs still hit Bash damage first (note that these two live cohorts ran separately under an earlier code grader, so they do not isolate the causal effect of adding Bash).

\section{Small Updates Can Disappear in Lower Precision}
\label{sec:precision}

Moving to gentler edits exposed another failure mode: exporting weights to lower precision can erase small updates. Training loops accumulate updates in fp32, whereas released checkpoints are saved and served in bf16 (7 explicit mantissa bits). Under round-to-nearest casting \citep{sahai2026lost}, an update $\Delta w$ away from zero from a normal bf16 weight $w_0$ snaps back to $w_0$ whenever it stays below half a bf16 unit in the last place:
\begin{equation}
  \mathrm{bf16}(w_0 + \Delta w) = w_0 \quad\text{whenever}\quad
  |\Delta w| < \frac{1}{2}\,\mathrm{ULP}_{\mathrm{bf16}}(w_0) = 2^{\lfloor \log_2 |w_0| \rfloor - 8}.
  \label{eq:bf16_ulp}
\end{equation}
Because $\frac{1}{2}\,\mathrm{ULP}_{\mathrm{bf16}}(w_0) / |w_0| \in [2^{-9}, 2^{-8}]$, any parameter change smaller than roughly $0.2\%\text{--}0.4\%$ of $|w_0|$ vanishes upon export (for $w_0 \approx 0.0123$, $\mathrm{ULP}_{\mathrm{bf16}}(w_0) = 2^{-14} \approx 6.10 \times 10^{-5}$).

\begin{figure}[t]
  \centering
  \includegraphics[width=0.90\textwidth]{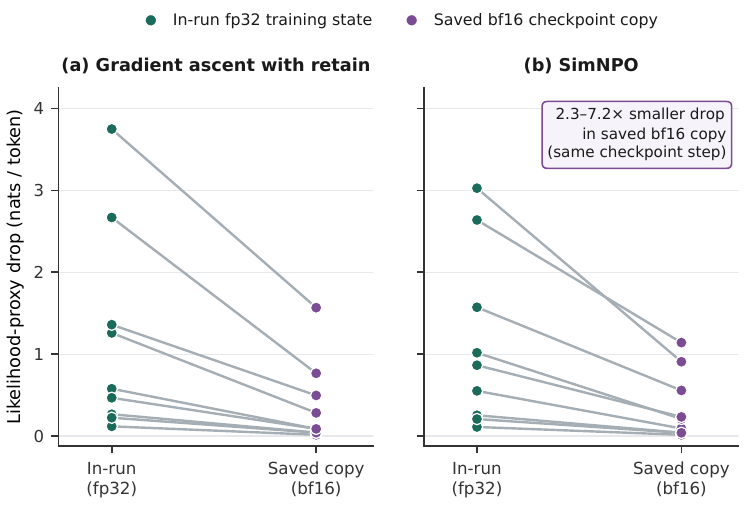}
  \caption{Comparison of in-run (fp32) and saved-copy (bf16) likelihood-proxy drops on verified Perl solutions in Qwen3-8B (nats per token; single seed, 18 checkpoints up to step 50 of 150) for (a)~gradient ascent with a retain term and (b)~SimNPO. Saved bf16 copies showed likelihood drops 2.3--7.2$\times$ smaller than the fp32 training states from which they were saved. Because saved-copy evaluation changes both stored weight precision and inference arithmetic, this measurement does not isolate storage rounding alone, and reflects token likelihood rather than code execution pass rates.}
  \label{fig:saved_copy}
\end{figure}

Prior work reported related precision effects across formats: 4-bit quantization restores unlearned information \citep{zhang2025catastrophic}, fp16-to-bf16 conversion weakens forgetting in gradient ascent and NPO under ReLearn \citep{xu2025relearn}, and bf16 export reverts low-learning-rate fine-tuning changes in ternary models \citep{sahai2026lost}. In our own initial Qwen3-8B Perl runs using gradient ascent with a retain term and Simple Negative Preference Optimization (SimNPO; \citealp{fan2025simplicity}), whose reference-free objective optimizes mean per-token forget NLL $\bar{\ell}_\theta(y \mid x)$ with an averaged neighbor and general-text retain term,
\begin{equation}
  \mathcal{L}_{\mathrm{SimNPO}}(\theta) = -\frac{2}{\beta}\,\mathbb{E}_{(x,y)\sim \mathcal{D}_f}\!\left[\log \sigma\!\Bigl(\beta\bigl(\bar{\ell}_\theta(y \mid x) - \delta\bigr)\Bigr)\right] + \mathcal{L}_{\mathrm{retain}}(\theta)
  \label{eq:simnpo}
\end{equation}
($\beta = 0.1, \delta = 0.0$), saved bf16 checkpoints showed token-likelihood drops 2.3--7.2$\times$ smaller than the fp32 training states we saved them from (Figure~\ref{fig:saved_copy}). A follow-up removal-side precision test across four methods on Perl and film facts was inconclusive overall because the RMU film control barely moved and several comparisons stayed unresolved.

The precision effect stood out more clearly on recovery attacks. Starting from a narrow-RMU Perl-unlearned Qwen3-8B checkpoint, we ran full-model fp32 relearning across a 20-setting grid (5 learning rates from $10^{-6}$ to $10^{-4}$ by 4 example budgets from 8 to 512, 3 seeds each) and graded execution on 75 test Perl problems. For each attack seed, we measured the share of fp32 recovery gain lost under 16-bit evaluation, $\Lambda_{\mathrm{lost}} = 1 - (A_{\mathrm{bf16}} - R_{\mathrm{bf16}})/(A_{\mathrm{fp32}} - R_{\mathrm{fp32}})$ (where $A$ and $R$ are attacked and removed-model pass rates in each precision), and averaged across the three seeds. Across the 13 settings that met our pre-registered recovery-eligibility bar in every seed, the fraction of changed weight values rounded away in bf16 correlated strongly with $\Lambda_{\mathrm{lost}}$ (Spearman $\rho = 0.85$).

\begin{figure}[!tp]
  \centering
  \includegraphics[width=0.90\textwidth]{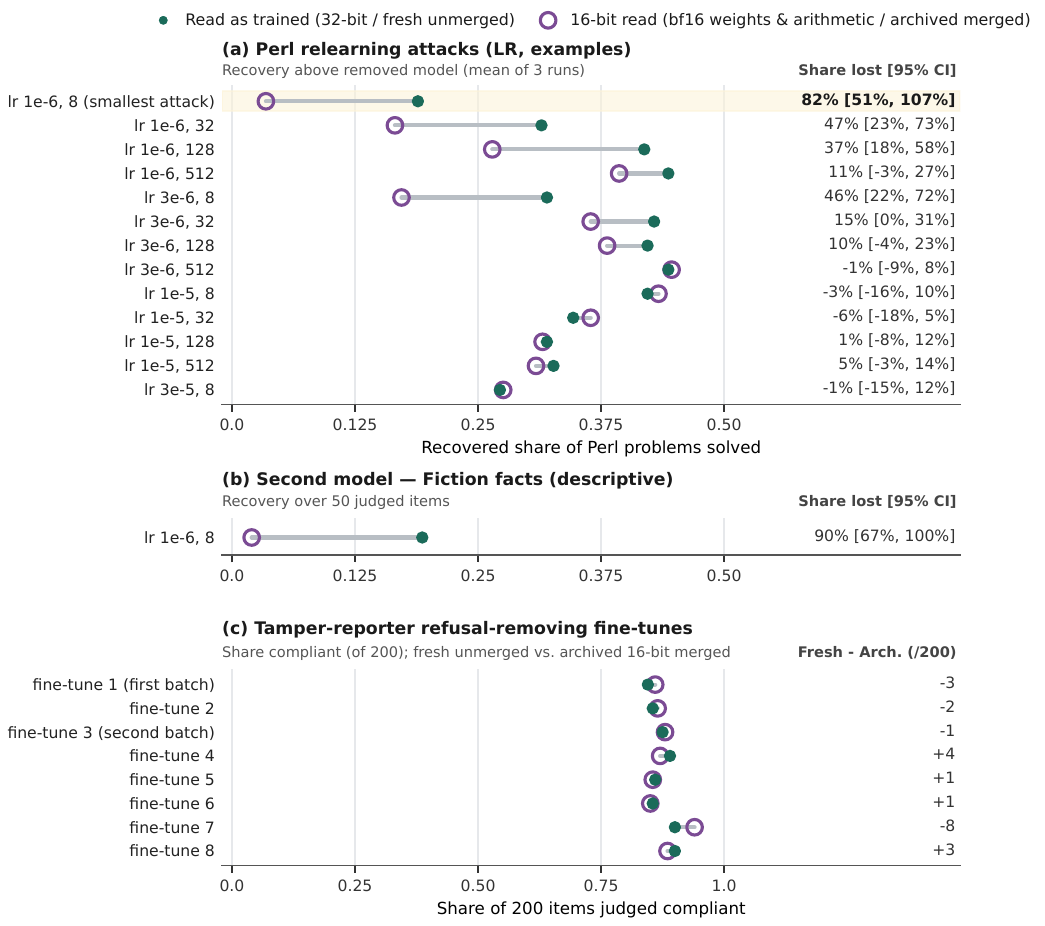}
  \caption{Recovery and compliance measured in training precision (filled teal dots) versus 16-bit reads (hollow purple rings). (a)~Perl relearning attacks on a narrow-RMU Qwen3-8B checkpoint across 13 eligible of 20 tested grid settings, measuring working programs on 75 test problems (mean of three runs; 95\% CIs use an unclipped item-level bootstrap over the 75 problems, not new models or repeats, so values exceed 100\% when the 16-bit pass rate dips slightly below the removed-model baseline on resampled subsets). Eligibility requires measurable recovery in every repeat; settings at $\mathrm{lr}=3\times 10^{-5}$ (32, 128, 512 examples) and $\mathrm{lr}=10^{-4}$ (all budgets) did not qualify. (b)~Descriptive comparison on Harry Potter facts in an RMU-edited Gemma-2-2b-it checkpoint (50 judged items). (c)~Refusal-removing LoRA fine-tunes on Qwen2.5-1.5B-Instruct tamper-reporting models (share of 200 draws judged compliant), comparing fresh unmerged reads against archived 16-bit merged checkpoints. Because weight precision and serving arithmetic changed together (and panel~(c) also changes read form), comparisons belong within each row rather than across panels.}
  \label{fig:precision_recovery}
\end{figure}

\begin{figure}[!tb]
  \centering
  \includegraphics[width=0.95\textwidth]{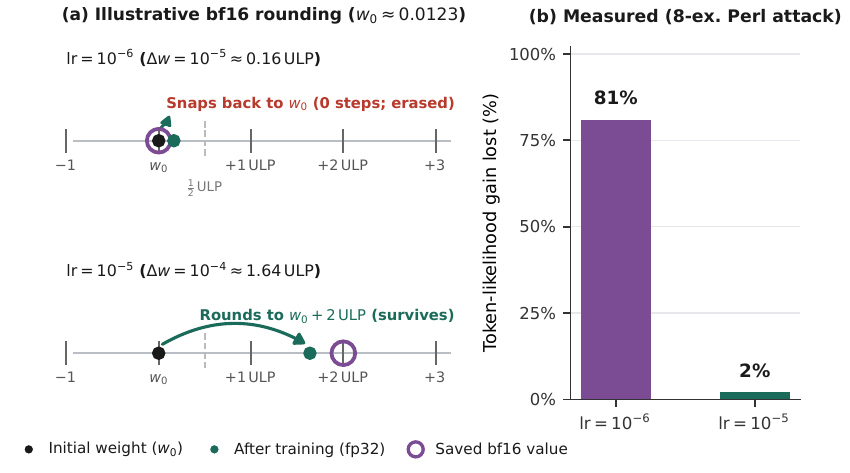}
  \caption{(a)~Illustrative bf16 round-to-nearest-even behavior around $w_0 = \mathrm{bf16}(0.0123)$ under a synthetic update rule ($\Delta w = 10 \times \mathrm{lr}$, approximating ten Adam steps). Tick marks show exact neighboring bf16 values spaced by $1\,\mathrm{ULP} \approx 6.10 \times 10^{-5}$. At $\mathrm{lr}=10^{-6}$, the fp32 update ($\approx 0.16\,\mathrm{ULP}$) falls below the half-ULP boundary and snaps back to $w_0$ when cast to bf16; at $\mathrm{lr}=10^{-5}$ ($\approx 1.64\,\mathrm{ULP}$), the update crosses into a new bf16 bin and survives. (b)~Measured share of \emph{token-likelihood} recovery lost in the 16-bit read (changing both stored precision and serving arithmetic, unlike storage rounding alone in panel~(a) and execution recovery in Figure~\ref{fig:precision_recovery}) for an 8-example relearning attack on a Perl-removed Qwen3-8B checkpoint: 81\% lost at $\mathrm{lr}=10^{-6}$ versus 2\% at $\mathrm{lr}=10^{-5}$.}
  \label{fig:bf16_rounding}
\end{figure}

Recovery loss concentrated in the gentlest attacks (Figure~\ref{fig:precision_recovery}a). At 8 examples and $\mathrm{lr} = 10^{-6}$, $\Lambda_{\mathrm{lost}} = 82\%$ (95\% CI: $[51\%, 107\%]$; the unclipped item-bootstrap upper bound exceeds 100\% whenever $A_{\mathrm{bf16}} < R_{\mathrm{bf16}}$ on resampled problem subsets). A descriptive check on Harry Potter facts in an RMU-edited Gemma-2-2b-it checkpoint (50 judged items, 8 examples at $\mathrm{lr} = 10^{-6}$) showed a similar 90\% loss (Figure~\ref{fig:precision_recovery}b). At learning rates of $10^{-5}$ and above, eligible settings retained almost all measured recovery, so bf16 casting is not a defense. Figure~\ref{fig:bf16_rounding} illustrates how sub-half-ULP updates snap back on bf16 rounding alongside the measured token-likelihood recovery loss on an 8-example Perl attack (81\% lost at $\mathrm{lr}=10^{-6}$ vs.\ 2\% at $\mathrm{lr}=10^{-5}$).

Still, counting rounded weights does not predict behavior on its own. We tested eight refusal-removing LoRA fine-tunes on Qwen2.5-1.5B-Instruct tamper-reporting models, which are fine-tuned to emit a standardized disclosure note when weight edits suppress refusal (Figure~\ref{fig:precision_recovery}c). Merging those adapters into bf16 wiped out 37--41\% of intended nonzero fp32 update entries, yet compliance counts on 200 harmful-request draws (graded by Qwen2.5-32B-Instruct) shifted by only $-8$ to $+4$ items and none of the pre-registered stage verdicts or readout count bands changed. Weight diffs can mislead in either direction: gentle updates can vanish on bf16 cast, or large shares of vanished entries can leave behavior intact. Reliable evaluation requires testing the exact weight precision and serving setup an attacker or user will run \citep{zhang2025catastrophic, sahai2026lost}.

\FloatBarrier

\section{Weight and Feature Localization Did Not Beat Untargeted Edits}
\label{sec:localization}

Even in exact serving precision, untargeted edits stay caught between sharp cliffs and fragile updates because target knowledge and benign skills share the same layers. Isolating the specific parameters or features behind a concept first should, in principle, let us delete the target without hurting unrelated behavior \citep{guo2024mechanistic, farrell2024applying}. We tested four localization methods (Table~\ref{tab:localization}).

\paragraph{Adversarial Parameter Decomposition (VPD).}
Using Goodfire's adVersarial Parameter Decomposition (VPD; \citealp{bushnaq2026interpreting}) on Goodfire's 67M model, we added a deletability objective rewarding components whose removal suppressed a target fact without hurting other behavior across 34 geography and world-knowledge training facts (6 settings, 204 cells, 1 fit seed). Even in-sample on those training facts, no setting worked: weak edits left targets intact, stronger edits damaged neighboring facts or general text, and different facts sometimes selected identical components. Because optimization failed in-sample, we did not run held-out deletion or relearning attacks.

\paragraph{Targeted Parameter Decomposition (tPD).}
On film-director facts in Qwen3-1.7B \citep{yang2025qwen3}, targeted Parameter Decomposition (tPD; \citealp{vigouroux2026targeted}) decomposed only a subset of MLP layers (layers 18--27) at usable reconstruction quality. Deleting localized components showed a specificity advantage over random pieces (clearly on one decomposition seed, slightly on the other), but lowered accuracy by only 0.20--0.24 from a 0.90 baseline (to 0.66--0.70) before damaging neighboring film facts. Untargeted RMU removed roughly 0.70 accuracy points (not gate-tested at that depth). We skipped recovery attacks on tPD because a failed attack is only informative when it can undo a strong baseline \citep{guo2024mechanistic}, and our attacks failed to recover half of the performance the RMU control removed ($(\mathrm{acc}_{\mathrm{att}} - \mathrm{acc}_{\mathrm{rem}}) / (\mathrm{acc}_0 - \mathrm{acc}_{\mathrm{rem}}) < 0.50$).

\paragraph{Feature-guided editing (PISCES).}
Under PISCES \citep{gurarieh2025precise} on Gemma-2-2b-it (targeting Harry Potter and Republic of Ireland concepts), our recovery attacks restored most matched RMU models while leaving PISCES models below our recovery threshold. Applying the same PISCES edit operator to randomly chosen features, however, also left several random controls below the recovery threshold despite weaker initial removal. Edited PISCES models passed a fresh-learning check on newly taught invented facts (5 of 8 vs.\ 6 of 8 for the base model), which rules out total learning collapse without showing broader usefulness survived. Because random controls were incomplete and not fully damage-matched, the evaluation remained insufficient to separate feature localization from the edit operator.

\paragraph{Entity-token embedding resets.}
In a matched Pile retrain of Goodfire's 67M model, we reset 140 tied entity-token embedding rows covering 27 world-knowledge targets, which dropped held-out recall to 0.06. To test whether donor embeddings could restore those facts, we fitted an unconstrained centered affine least-squares map in fp64 over $N = 41{,}992$ non-reset anchor tokens ($\mathbf{E}_d, \mathbf{E}_r \in \mathbb{R}^{N \times d}$ with row means $\boldsymbol{\mu}_d, \boldsymbol{\mu}_r \in \mathbb{R}^{1 \times d}$, $d = 768$) and transplanted only the 140 reset rows:
\begin{equation}
  \mathbf{M}^\star = \arg\min_{\mathbf{M} \in \mathbb{R}^{d \times d}} \left\|(\mathbf{E}_d - \mathbf{1}\boldsymbol{\mu}_d)\mathbf{M} - (\mathbf{E}_r - \mathbf{1}\boldsymbol{\mu}_r)\right\|_F^2, \qquad
  \tilde{\mathbf{e}}_i = (\mathbf{e}_{d,i} - \boldsymbol{\mu}_d)\mathbf{M}^\star + \boldsymbol{\mu}_r.
  \label{eq:donor_ols}
\end{equation}
Transplanting from Goodfire's released 67M Pile checkpoint restored recall to 0.82 (vs.\ 0.41 under centered scaled orthogonal Procrustes), and transplanting from a twin donor (an independent Pile retrain of the same architecture using a second data seed) restored recall to 0.77 (vs.\ 0.29 under Procrustes). Because donor embeddings supply factual associations of their own (lifting an entity-filtered cleanroom reference from 0.03 to 0.60), this recovery leaves open how much knowledge lived in the embedding rows versus deeper layers. In addition, the approach depends on dedicated entity tokens that do not transfer to shared BPE vocabularies such as Qwen3 (where only 1 of 307 checked tokens qualified).

\begin{table}[!tb]
  \centering
  \small
  \caption{Summary of our four parameter- and feature-localization experiments.}
  \label{tab:localization}
  \begin{tabularx}{\textwidth}{>{\RaggedRight\arraybackslash}p{0.22\textwidth} >{\RaggedRight\arraybackslash}p{0.34\textwidth} L}
    \toprule
    \textbf{Localization test} & \textbf{Setup and scope} & \textbf{Outcome} \\
    \midrule
    VPD with deletability objective &
      Goodfire 67M \citep{bushnaq2026interpreting}; 34 world-knowledge training facts $\times$ 6 settings (204 cells), 1 fit seed &
      In-sample optimization failure; weak edits left targets intact, stronger edits damaged neighboring facts or general text. \\
    \addlinespace[5pt]
    tPD film deletion &
      Qwen3-1.7B \citep{yang2025qwen3}; 2 decompositions, MLP layers 18--27 &
      Lowered accuracy by 0.20--0.24 from 0.90 before neighbor damage vs.\ $\sim$0.70 for untargeted RMU (not gate-tested at that depth); skipped attacks because they failed to recover half of what the RMU control removed. \\
    \addlinespace[5pt]
    PISCES feature-guided editing &
      Gemma-2-2b-it (Harry Potter, Republic of Ireland); matched RMU and random-feature controls &
      Inconclusive (insufficient): resisted recovery better than matched RMU, but several random-feature controls also resisted recovery despite weaker initial removal. \\
    \addlinespace[5pt]
    Entity-token embedding resets &
      Goodfire 67M matched Pile retrain; reset 140 tied entity rows and restored via least-squares mapped donor rows &
      Dropped recall to 0.06 and restored it to 0.77--0.82 via donor rows, but donor embeddings carry their own factual associations; dedicated tokens do not transfer to shared BPE vocabularies (Qwen3: 1/307). \\
    \bottomrule
  \end{tabularx}
\end{table}

Across all four experiments, localizing first gave no clear win over untargeted edits. A localization method needs to beat untargeted baselines on the removal-versus-damage frontier and withstand calibrated recovery attacks that outscore matched random-site controls \citep{lee2025does, guo2024mechanistic}.

\FloatBarrier

\section{Discussion, Limitations, and Future Directions}
\label{sec:future}

Across our experiments, post-hoc unlearning struggled whenever target knowledge and benign skills shared the same weights. Pushing edits deeper triggered sharp capability cliffs and collateral damage to neighboring tasks; keeping edits gentle pushed them near bf16 rounding boundaries and left answers accessible under other query formats or small fine-tuning attacks. Post-hoc unlearning cannot carry an open-weight safety case on its own today.

\paragraph{Limitations.}
Every experiment here uses benign proxies (facts and Perl programming) on 67M to 8B public models rather than real CBRN workflows at frontier scale. Several results come from exploratory development grids or shared selection sets (detailed experiment-by-experiment in Appendix~\ref{app:evidence}), and every failure to recover a capability is bounded by the finite attack budgets and query formats we ran.

\paragraph{Future directions.}
Next, we plan to test three approaches:
\begin{enumerate}[leftmargin=1.8em, itemsep=4pt, topsep=4pt]
  \item \textbf{Stronger post-hoc erasure and entanglement:} Pairing deeper weight edits with targeted repair (such as resetting activation subspaces and distilling benign behavior back from a filtered teacher) may raise recovery cost without broad collateral damage. Binding refusal directly into task-critical MLP computations could also force an attacker who strips refusal to degrade the underlying capability.
  \item \textbf{Disrupting multi-step workflows:} Dangerous biosecurity or cyber capabilities chain many steps together. Breaking the transitions that connect steps may prove more selective than erasing individual dual-use facts needed for benign work.
  \item \textbf{Training-time modularity:} Post-hoc localization struggled partly because pretraining scatters related concepts across shared parameters. Routing target-domain gradients into dedicated, removable parameter subsets during training could make surgical deletion practical before weights freeze.
\end{enumerate}

Open-weight models have not yet reached the capability thresholds where weight-level removal is critical, which makes now the right time to test these defenses. We ran this project as an empirical sweep of existing unlearning methods, many originally built for privacy or memorization benchmarks rather than adversarial capability removal. None of the tested methods met our criteria. We will keep testing stronger approaches ahead of that capability curve so that, when open models do cross high-risk thresholds, we have tools sharp enough to preserve dual-use utility while cutting off severe misuse.

{\small
\bibliography{references}
}

\clearpage
\appendix

\section{Models Evaluated}
\label{app:models}

The experiments reported in this paper used Qwen3-8B, Qwen3-4B, Qwen3-1.7B, Qwen2.5-1.5B-Instruct, Gemma-2-2b-it (with Gemma Scope sparse autoencoders), and Goodfire's 67M model (alongside a compatible 67M Pile-trained donor checkpoint and a matched Pile retrain), including edited and fine-tuned versions of those models, as well as Qwen3-0.6B to generate candidate answers and Qwen2.5-32B-Instruct for compliance grading. No additional models were used.

\section{Summary of Experimental Evidence}
\label{app:evidence}

Table~\ref{tab:evidence} summarizes each experiment reported in the main text together with its primary findings and methodological caveats.

\begingroup
\small
\setlength{\LTpre}{6pt}
\setlength{\LTpost}{6pt}
\begin{longtable}{>{\RaggedRight\arraybackslash}p{0.25\textwidth} >{\RaggedRight\arraybackslash}p{0.68\textwidth}}
  \caption{Evidence at a glance across all experiments, including key scope limits and caveats.}
  \label{tab:evidence} \\
  \toprule
  \textbf{Experiment} & \textbf{Key findings and methodological caveats} \\
  \midrule
  \endfirsthead

  \multicolumn{2}{l}{\footnotesize\itshape Table~\ref{tab:evidence} (continued)} \\
  \toprule
  \textbf{Experiment} & \textbf{Key findings and methodological caveats} \\
  \midrule
  \endhead

  \midrule
  \multicolumn{2}{r}{\footnotesize\itshape Continued on next page} \\
  \endfoot

  \bottomrule
  \endlastfoot

  Geography decoys &
    Models recognized answers they refused to give when the candidate list included the true answer. Some evaluation facts were used during scoping. \\
  \addlinespace[4pt]

  Deep RMU edits &
    In the deepest edits, ranking candidates by Yes probability still found every true answer despite worse completion ranking. Every configuration exceeded the limits on damage to other behavior. \\
  \addlinespace[4pt]

  Attacking DDO refusal &
    Refusal fell under attack while recognition stayed strong and generation remained impaired. Recognition constrained attack selection; generation used different items and was not the attack's objective. \\
  \addlinespace[4pt]

  All-layer RMU skill removal &
    Perl performance eroded, then collapsed sharply in the measured run. Lost execution performance does not prove the skill was deleted. \\
  \addlinespace[4pt]

  RepNoise skill removal &
    All fixed repeats showed positive one-step losses of working Perl, with later Bash damage estimates above the limit. The counting rule followed initial results; intervals describe individual runs on shared selection problems, not a general cliff frequency. \\
  \addlinespace[4pt]

  Layer placement &
    The chosen single-layer edit met the measured stopping limits in every repeat (selected in hindsight, and all five recovered $\geq 50\%$ from 8 examples). Wider Perl edits rarely did; most multi-layer film edits stopped cleanly later. Recipes, placements, and read schedules limit the comparison. The pilot illustrates trajectories, not their frequency. \\
  \addlinespace[4pt]

  Repeated probability checks &
    Every run reached its assigned threshold on the selection problems. RepNoise crossings clustered closely; RMU crossings varied. Different thresholds and arithmetic prevent a comparison of computational cost; these are not graded skill measurements. \\
  \addlinespace[4pt]

  NPO near its stopping threshold &
    Across the full fixed cohort, losses stayed shallow at the borrowed threshold on problems used to choose the edit. We resolved only a small rise; unresolved changes do not establish zero change or exclude cliffs elsewhere. \\
  \addlinespace[4pt]

  Gentle removal after casting &
    The test was inconclusive: the film control barely moved and some comparisons remained unresolved. Measurable removals had smaller estimated effects at the gentlest rate; arithmetic changed too. \\
  \addlinespace[4pt]

  Attack recovery after casting &
    Execution recovery varied with the attack setting, with the largest losses in small, gentle attacks. Arithmetic changed too; only settings with measurable recovery in every repeat qualified, and excluded settings say nothing about retention. \\
  \addlinespace[4pt]

  VPD fact deletion &
    No tested edit suppressed its target while preserving neighboring facts and general-text performance. We did not test held-out deletion or relearning. \\
  \addlinespace[4pt]

  Targeted parameter decomposition &
    Within the covered layers, specificity gains ranged from slight to clear, without improving the removal--damage tradeoff. A weak recovery control prevented meaningful attack testing. \\
  \addlinespace[4pt]

  PISCES feature-located edits &
    Under the attacks tried, PISCES edits and some random controls stayed below the recovery threshold; a matched control topic reached it. Incomplete controls and damage to other abilities left localization's benefit unresolved (insufficient). \\
  \addlinespace[4pt]

  Refusal-training recipes &
    Training with ablation preserved more refusal against related attacks. Both recipes yielded to fine-tuning at the smallest measured budget; smaller budgets remain unresolved. \\
  \addlinespace[4pt]

  Fixed stopping rule &
    On all-layer RMU, runs often hit collateral damage before reaching the target (3 of 5 runs). These observations do not establish a reliable way to choose a selective stopping point. \\
  \addlinespace[4pt]

  Adding Bash to retention training &
    Every fresh all-layer RMU run still encountered damage before the stop. The cohorts differed, so the comparison does not isolate the effect of adding Bash. \\
  \addlinespace[4pt]

  Reporter re-reads &
    Across the declared comparisons, many update entries disappeared without changing reported decisions. These comparisons mixed fresh and archived reads, so they do not isolate precision's contribution. \\
  \addlinespace[4pt]

  Entity-token resets &
    Resetting rows suppressed recall; donor rows restored much of it. The donor could add information, leaving storage location unresolved; we did not measure usefulness after the transplant. \\
\end{longtable}
\endgroup

\end{document}